\documentclass[sigconf]{acmart}

\setcopyright{cc}
\setcctype[4.0]{by}
\acmDOI{10.1145/3858063.3858863}
\acmISBN{979-8-4007-3087-0/2026/11}
\startPage{1547}

\acmConference[ICAIF '26]{7th ACM International Conference on AI in
  Finance}{November 14--17, 2026}{Milan, Italy}
\acmBooktitle{7th ACM International Conference on AI in Finance (ICAIF '26), November 14--17, 2026, Milan, Italy}
\acmYear{2026}
\copyrightyear{2026}

\newcommand{\eps}{\varepsilon}
\newcommand{\BR}{\mathrm{BR}}
\newcommand{\Neff}{N_{\mathrm{eff}}}
\newcommand{\hSP}{h_{\mathrm{SP}}}
\newcommand{\hPO}{h_{\mathrm{PO}}}
\newcommand{\gPO}{\gamma_{\mathrm{PO}}}
\newcommand{\geff}{\gamma_{\mathrm{eff}}}

\newcommand{\Kdb}{K_{\mathrm{db}}}
\newcommand{\Kmax}{K_{\max}}
\newcommand{\hRef}{h_{\mathrm{ref}}}  
\newcommand{\gbar}{\bar{g}}
\newcommand{\Wone}{W_1}
\newcommand{\code}[1]{\texttt{#1}}

\DeclareCaptionFont{captionmath}{\boldmath}
\begin{document}

\title[REFLEX: Analytic Stability Boundaries for Performative Market Making]%
{REFLEX: Reflexive Equilibrium Fixed-point Learning for Endogenous
eXchanges}

\author{Vignesh Nagarajan}
\authornote{Corresponding author.}
\authornote{Both authors contributed equally to this research.}
\affiliation{%
  \department{Department of Computer Science \& Engineering}
  \institution{Texas A\&M University}
  \city{College Station}
  \state{Texas}
  \country{USA}}
\email{nrvignesh08@gmail.com}
\orcid{0009-0007-8816-8359}

\author{Shriraghav Ashok}
\authornotemark[2]
\affiliation{%
  \department{Haas School of Business}
  \institution{University of California, Berkeley}
  \city{Berkeley}
  \state{California}
  \country{USA}}
\email{shriraghav.a524@gmail.com}
\orcid{0009-0008-8456-9697}

\begin{abstract}
In over-the-counter corporate bond markets, dealers compete for client trades by
quoting bid and ask prices. Tighter quotes attract more business, but also
informed customers more likely to trade ahead of adverse price moves, leaving
the dealer holding the risk. As dealers increasingly use machine
learning to set quotes, they retrain these models on the trades their own quotes
attract, creating a feedback loop in which each model reshapes the market that
generates its next training data. The question is therefore not only whether a
quoting model performs well, but whether the market it creates stays stable as
the model learns from it. Existing performative prediction theory gives a sharp
stability condition, yet expresses it through abstract properties of the
learning objective a trading desk cannot measure before deployment. We
introduce \textsc{Reflex}, a framework that replaces those unobservable
quantities with three measurable features of dealer behavior: how strongly
trading volume responds to tighter quotes, how sharply the dealer's objective
bends around its optimum, and how quickly informed flow increases as spreads
narrow. \textsc{Reflex} combines these into a single retraining modulus,
a pre-deployment stability margin estimated from a desk's own quote and
execution history that predicts whether repeated retraining will converge or
amplify itself. In simulation, predicted and measured moduli agree within
$8\%$ where the loop contracts, though the uncorrected instability threshold
is optimistic ($4.70$ predicted, $3.17$ measured). Competing dealers increase instability by $1.74\times$ with two
and $3.16\times$ with three, against predicted $2\times$ and $3\times$. Where
ordinary retraining becomes
unstable at modulus $1.21$, a structurally anchored correction converges as
blind retraining collapses. Calibrated over $36$ years of public market data,
stability headroom falls roughly $4.4\times$ for investment grade and
$4.3\times$ for high yield from calm to crisis regimes.
\end{abstract}



\maketitle

\section{Introduction}

Corporate bonds trade over the counter. A client requests a quote, a dealer
posts a half-spread $h$, and the dealer's inventory absorbs the flow. Quoting
on these desks is increasingly algorithmic, and quoting policies are
increasingly fit to historical flow.

That flow is not exogenous. A tighter quote wins more volume, and it also
summons more informed, or ``toxic,'' flow that picks the dealer off before
adverse price moves. A wider quote starves both channels. The quoting policy
$\phi$ therefore induces the flow distribution $D(\phi)$ it will next be
trained on. Fitting on the induced flow, redeploying, and refitting has the
structure of \emph{repeated risk minimization} (RRM) in performative
prediction \cite{perdomo2020performative}. There, RRM converges to a
performatively stable point whenever $\eps<\gamma/\beta$, where $\eps$ bounds
the sensitivity of the distribution map and $\gamma,\beta$ are the strong
convexity and smoothness of the loss.

The theorem is sharp, but its constants are Lipschitz \emph{assumptions}. A
desk cannot evaluate $\eps$, $\beta$, or $\gamma$ before deploying, so it
cannot know whether its own retraining cadence is stable. Similarly, a
supervisor watching many dealers retrain against one shared pool of informed
flow cannot know whether competition among learners is manufacturing systemic
fragility. Post-trade TRACE reporting reveals the spreads that result, not
the retraining loop that produced them. Both questions are quantitative, and
answering either needs the constants themselves. We compute a market-level
stability margin for algorithmic liquidity provision, which puts the problem
in the AI-governance and systemic-risk scope of AI in finance.

To address these limitations, this paper introduces REFLEX
(\emph{Reflexive Equilibrium Fixed-point Learning for Endogenous
eXchanges}), which builds that retraining loop \emph{inside} a
structural OTC market-making model in the
Gu\'eant--Lehalle--Fern\'andez-Tapia (GLFT) family. There the loop is a
scalar cobweb map, and its local counterparts of the three constants follow
from microstructure primitives. We check every closed form against a learned
simulator loop. We contribute six closed-form results, each tested by
experiment:
\begin{enumerate}
\item[\textbf{R1}] \emph{An analytic stability boundary.} We give the
  cobweb's $\gamma$, $\beta$, and $\eps$ in closed form from fill-curve
  curvature, P\&L scale, and toxic-flow slope, so the modulus
  $m=\eps\beta/\gamma$ is known before deployment. Measured moduli track it
  within $8\%$ where the loop contracts.
\item[\textbf{R2}] \emph{A performative-gradient (PerfGD) correction},
  closed-form and estimated. One extra scalar per step converges past the
  RRM boundary, and its structurally anchored learned counterpart settles at
  the realized performative optimum where blind retraining collapses.
\item[\textbf{R3}] \emph{A multi-dealer boundary} $\eps<\gamma/(\Neff\beta)$.
  Competition destabilizes the market a factor $\Neff$ before any single
  dealer would, and a genuine shared-pool market measures $1.74\times$ and
  $3.16\times$ amplification at $N=2,3$.
\item[\textbf{R4}] \emph{Finite-sample robust certificates.} An
  $O(1/\sqrt{n})$ ambiguity radius bought by common random numbers (CRN)
  issues stable, unstable, or \emph{undecided} verdicts, separating
  statistical from structural uncertainty.
\item[\textbf{R5}] \emph{Factor-model scaling.} The $d{\times}d$ modulus
  matrix has spectral boundary $\rho(M)<1$ at $O(dk^2)$ cost, and
  $\rho(M)$ is flat from $8$ to $128$ bonds on our calibration.
\item[\textbf{R6}] \emph{Lazy deployment.} The $K$-step retraining map
  $\mu(K)=-m+c^K(1+m)$ yields deadbeat and maximal-stability cadences in
  closed form. With one fitted parameter, both match the measured probe for
  $K\le8$.
\end{enumerate}

The system side contributes an \emph{un-blinded} learned market-response
operator whose training identifies $dD/d\phi$. It adds three complementary
instruments for $\eps$: a CRN best-response probe, an entropic
optimal-transport estimate \cite{cuturi2013sinkhorn,feydy2019interpolating,
peyre2019computational}, and a fitted informed-flow curve
\cite{cont2014price}. It calibrates the simulator to 36 years of public
market data \cite{dickerson2023priced,friewald2012illiquidity}, and it builds
a verification layer of 66 numerical proof certificates plus Lean~4
\cite{moura2021lean,mathlib2020} formal skeletons. Negative results and
proxy-level data provenance are reported as such throughout.

\section{Related Work}
\label{sec:related}

Perdomo et al.\ \cite{perdomo2020performative} established the convergence of
RRM and its stochastic variants \cite{mendlerdunner2020stochastic}, and later
work estimates the distribution response $dD/d\phi$ to optimize performative
risk directly \cite{izzo2021learn,miller2021echo}. The framework now reaches
decision-dependent stochastic optimization \cite{drusvyatskiy2023stochastic},
distributions that carry their own dynamics
\cite{li2022state,brown2022stateful}, bandit feedback
\cite{jagadeesan2022regret}, multi-agent decision-dependent games
\cite{narang2022multiplayer}, and robust formulations; see
\cite{hardt2023performative} for a survey. In all of it,
$(\eps,\beta,\gamma)$ stay abstract constants of an unspecified loss.

On the market-making side, Avellaneda and Stoikov
\cite{avellaneda2008highfrequency} and Gu\'eant et al.\
\cite{gueant2013dealing} derived closed-form optimal quotes under exponential
fill intensities, and later work added multi-asset dimensionality reduction
\cite{bergault2021size}, adverse selection and price reading
\cite{barzykin2025adverse}, the stochastic-control toolkit
\cite{cartea2015algorithmic}, and the empirical anatomy of flow--price
interaction \cite{cont2014price}. These models take the flow response to
quoting as a fixed input, so the learning loop that re-estimates the market
from its own induced data falls outside their scope.

To our knowledge, no prior work computes a performative stability modulus
from market structure.

\section{Market Model and the Retraining Loop}
\label{sec:model}

One deployment is one period of quoting by a dealer posting half-spread $h$.
Skew is zero for the scalar theory, and the multi-bond lift is R5. With
$\rho$ the latent liquidity ratio, benign notional follows the GLFT
exponential fill curve \cite{avellaneda2008highfrequency,gueant2013dealing}
and informed notional a spread-gated channel:
\begin{equation}
U(h)=A\rho\,e^{-kh},\qquad
\tau(h)=\rho\,\gbar\big(I_b+\alpha f I\,e^{-c_t h}\big),
\label{eq:flows}
\end{equation}
where $A,k$ are the arrival rate and demand elasticity, $I_b,I$ the base and
feedback informed intensities, $\alpha$ adversarial informativeness, $f$ the
toxicity-feedback gain (our control variable), $c_t$ the toxic spread decay,
and $\gbar\in(0,1)$ the mean of the informed traders' $\tanh$ signal gate.
Informed flow trades with the signal, so each unit costs the dealer an
adverse-selection severity $\psi>0$, a state constant from the same Gaussian
gate integrals. The expected one-deployment objective, with P\&L scale $P$,
quoting-cost weight $w$, anchor $\hRef$, and an inventory-variance penalty
$R_q$ carried through its curvature $R_q''=\lambda_q\ge0$, is
\begin{equation}
J(h;T)=P\big[\,hA\rho e^{-kh}+hT-\psi T-w(h-\hRef)^2-R_q(h)\,\big].
\label{eq:objective}
\end{equation}
The frozen toxic level $T$ is what makes the learner structurally blind.
Within a deployment the fitted model conditions on the deployed regime, so
the dealer best-responds to $T=\tau(h_{\mathrm{dep}})$ while only the benign
channel varies with the candidate $h$. Retraining is therefore a cobweb map,
\begin{equation}
h_{t+1}=\BR(h_t):=\arg\max_h J\big(h;\tau(h_t)\big),\qquad
m:=\big|\BR'(h^*)\big|,
\label{eq:cobweb}
\end{equation}
which contracts locally to the performatively stable spread $\hSP$ if $m<1$
and oscillates away from it if $m>1$. The simulator realizes \eqref{eq:flows} to \eqref{eq:cobweb} with
multi-bond flow, an informed-flow saturation cap, and a liquidity field that
\emph{inflates with realized flow}. The frozen closed forms omit those
endogenous channels, and the measurements below make them visible.

\section{Closed-Form Stability Theory}
\label{sec:theory}

We state the objects the experiments test. All six follow from the
one-deployment objective \eqref{eq:objective} by differentiation and the
implicit-function theorem; only R1's argument is given in full.

\textbf{R1 (analytic boundary).} Differentiating \eqref{eq:objective} gives
the cobweb's counterparts of the three performative constants in closed form.
They are local values at $h$, not the uniform bounds of
\cite{perdomo2020performative}:
\begin{align}
\gamma&=P\big[2w+A\rho k e^{-kh}(2-kh)+\lambda_q\big],\qquad \beta=P,
\notag\\
\eps(h)&=\rho\,\gbar\,\alpha f I c_t\,e^{-c_t h}.
\label{eq:constants}
\end{align}
Curvature is fill-curve curvature plus quoting-cost stiffness, the coupling
$\beta$ is the P\&L scale, and $\eps$ is the toxic-flow slope $|d\tau/dh|$, a
mean shift that lower-bounds the Wasserstein sensitivity of
\cite{perdomo2020performative}. The
implicit-function theorem gives $\BR'=-\eps\beta/\gamma$, an oscillatory
cobweb, hence
\begin{equation}
m(h^*)=\frac{\eps\beta}{\gamma}
=\frac{\rho\gbar\alpha f I c_t\,e^{-c_t h^*}}
       {2w+A\rho k\,e^{-k h^*}(2-k h^*)+\lambda_q},
\label{eq:modulus}
\end{equation}
with $P$ cancelling and $h^*$ the self-consistent fixed point of
\eqref{eq:cobweb}. The loop is locally stable if $m<1$ and unstable if
$m>1$. Every term is evaluable before any loop is run, so the boundary is a
\emph{prediction}.

\emph{Derivation and regularity.} Subscripts $1$ and $T$ denote partial
derivatives in $h$ and $T$. Where $\gamma(h)>0$ on the capped quoting interval
the frozen-$T$ objective is $C^2$ and strictly concave,
$\partial_{11}J=-\gamma<0$, so its maximiser is interior and unique and
$\partial_1 J(h_{t+1};\tau(h_t))=0$ defines $\BR$ implicitly. Differentiating
in $h_t$ gives $\BR'=-\partial_{1T}J\,\tau'/\partial_{11}J$, and
\eqref{eq:objective}--\eqref{eq:flows} supply $\partial_{1T}J=\beta$ and
$\tau'=-\eps$, which is \eqref{eq:modulus}; local stability is then Banach
contraction near $h^*$. R3, R5 and R6 reuse this step with $\partial_{1T}J$ replaced by a
rank-one coupling, by $\Gamma^{-1}E$, and by the $K$-step warm-started map.

Two structural corollaries matter for honest measurement. First, adverse
selection ($\psi T$) is constant in $h$ within a deployment, so it sets the
profit level and the echo-chamber gap of R2 but drops out of the boundary.
Second, the modulus \emph{saturates} at the self-consistent fixed point:
widening spreads decay $\eps(h^*)$ exponentially, and at default-like
constants it never crosses $1$. This is defensive widening, so measured
instability describes the retraining map at the \emph{operating} spread, and
we report it that way.

\textbf{R2 (PerfGD correction).} The performative objective is
$\Phi(h)=J(h;\tau(h))$. Un-blinding adds one closed-form term to the blind
gradient $G(h)=\partial_1 J(h;\tau(h))$:
\begin{equation}
\Phi'(h)=G(h)+\Delta(h),\qquad
\Delta(h)=-\beta\,(h-\psi)\,\eps(h),
\label{eq:perfgd}
\end{equation}
where $\Delta=\partial_T J\cdot(d\tau/dh)$ is the distribution-response term,
supplied entirely by the R1 closed forms, so the corrected ascent costs one
scalar more per step than RRM \cite{izzo2021learn}. Its convergence is
governed not by the cobweb modulus but by the objective curvature at the
performative optimum $\hPO$,
\begin{equation}
\gPO=\gamma+\beta\eps\,(2+c_t\psi-c_t\hPO),
\label{eq:gpo}
\end{equation}
which stays positive where $m>1$, so the corrected loop converges beyond the
RRM boundary. The stable-versus-optimal separation, the ``echo-chamber gap''
of \cite{miller2021echo}, is
$\hSP-\hPO=\beta\eps(\hSP-\psi)/(\gamma+\beta\eps)=O(\eps)$ in decision
space and $O(\eps^2)$ in value.

\textbf{R3 (multi-dealer systemic risk).} $N$ dealers share one informed pool
with toxic spillover $\kappa\in[0,1]$, so flow displaced by one dealer's
widening lands on the others. The joint best-response Jacobian is a rank-one
common-mode coupling. Its differential eigenvalues carry $(1-\kappa)m_1$ and
its common mode $-m_1\Neff$ with $\Neff=1+\kappa(N-1)$, so the joint loop is
stable if and only if
\begin{equation}
\eps<\frac{\gamma}{\Neff\,\beta},
\qquad
N_c=1/m_1
\label{eq:multidealer}
\end{equation}
is the critical dealer count at which a market of individually stable dealers
($m_1<1$) turns systemically unstable. Competition manufactures a
synchronized cobweb, making fragility a market property rather than a desk
property. ``Systemic'' here means this market-level retraining instability;
we do not model losses or market failure.

\textbf{R4 (robust boundary).} With $n$ probe estimates of $\eps$ (sample
mean $\hat\eps_n$, std $s$), the certificate
\begin{equation}
\hat\eps_n+\delta_n<\frac{\gamma}{\beta},\qquad
\delta_n=\frac{z_{1-a}\,s}{\sqrt{n}},
\label{eq:robust}
\end{equation}
declares \emph{stable}; $\hat\eps_n-\delta_n>\gamma/\beta$ declares
\emph{unstable}; anything between is \emph{undecided}. The parametric
$O(1/\sqrt{n})$ radius is bought by the CRN pairing of the probe, since a
naive finite difference of independent runs achieves only $O(n^{-1/3})$ after
bias--variance balancing. Pinning a crossing needs
$n_{\mathrm{req}}=O(\Delta^{-2})$ in the distance $\Delta$ to the boundary,
so statistical uncertainty is separated from structural uncertainty by
construction.

\textbf{R5 (factor scaling).} For $d$ bonds, stability is governed by the
modulus matrix $M=\beta\,\Gamma^{-1}E$ with
$E=\mathrm{diag}(\eps_i)$ and $\Gamma$ the cross-bond curvature, and the loop
contracts if and only if $\rho(M)<1$. A $k$-factor covariance makes $\Gamma$
diagonal-plus-low-rank, so $\rho(M)$ costs $O(dk^2)$ by Woodbury, with
truncation error linear in the residual factor variance $\lambda_{k+1}(C)$.
Large universes therefore stay certifiable at practical cost.

\textbf{R6 (lazy deployment).} Suppose each deployment takes only $K$ inner
gradient steps on the frozen-$T$ objective (step $\eta_h$,
contraction $c=1-\eta_h\gamma$), warm-started from the deployed spread
instead of the exact best response. The outer map slope becomes
\begin{equation}
\mu(K)=-m+c^K(1+m),
\label{eq:lazy}
\end{equation}
a one-parameter interpolation from inertia at $\mu(0)=1$ to the exact cobweb
as $\mu\to-m$. Three consequences follow in closed form. A \emph{deadbeat}
cadence $\Kdb=\ln\!\big(m/(1+m)\big)/\ln c$ gives $\mu=0$ and one-shot
convergence. For $m>1$ a stability window $K\le\Kmax=
\ln\!\big((m-1)/(m+1)\big)/\ln c$ keeps an RRM-unstable market stable, the
quantitative form of the greedy-versus-lazy gap in
\cite{perdomo2020performative,mendlerdunner2020stochastic}. And an observer
who fits the plain boundary to a lazy loop reads a two-branch \emph{effective}
curvature $\geff(K)=\gamma\,m/|\mu(K)|$, under-estimating stiffness below the
equal-modulus cadence and over-estimating it above.

\section{Experimental Setup}
\label{sec:system}

\subsection{Learned Operator and the Four Loop Modes}
The simulator is a multi-bond OTC market with benign and informed clients,
inventory, an informed-flow saturation cap, and a latent liquidity field
inflated by realized flow. The learned market-response operator $T_\theta$
(a PyTorch MLP, two $\tanh$ layers of width 64 with distribution heads over
per-bond features, fit by Adam) predicts next-deployment flow
distributions conditioned on a per-deployment \emph{policy summary}, and is
fit over a sliding window of past deployments. With window $\ge2$ the
derivative of its prediction w.r.t.\ the summary, the learned $dD/d\phi$, is
identified by automatic differentiation; a single-window fit is structurally
blind and supplies the RRM baseline.

Four retraining modes close the loop. \textbf{(i)}~Blind RRM refits on the
latest induced flow. \textbf{(ii)}~\emph{PerfGD-analytic} uses the
closed-form $\Delta$ of \eqref{eq:perfgd} as a surrogate gradient.
\textbf{(iii)}~\emph{PerfGD-learned} consumes the free-form learned
$dD/d\phi$ and is reported as a negative result.
\textbf{(iv)}~\emph{PerfGD-structural} never asks the network for a
derivative. It instead fits the theory's own response families
\begin{equation}
\hat\tau(h)=\hat C_0+\hat C_1 e^{-\hat c h},\qquad
\hat u(h)=\hat A_u e^{-\hat k_u h},
\label{eq:structfit}
\end{equation}
plus the realized severity $\hat\psi$, to the loop's \emph{own} deployment
history (per-bond, per-step samples over a 12-deployment window; a small
quote jitter supplies the within-deployment identification the echo chamber
cannot destroy). It then ascends $\hat\Phi'=\hat G+\hat\Delta$ with step
$1/\hat\gPO$ under a $35\%$ relative step cap and an anti-echo
freeze that refuses to steer on unidentified fits. Its target is the
\emph{realized} performative optimum, judged against independent structural
fits (Sec.~\ref{sec:gap}). Both safeguards come from
failures we measured: exponential response fits on a narrow spread window are
line-degenerate and extrapolate badly, and the unconstrained early ascent
rails against its spread cap.

\subsection{Three Instruments for $\eps$ and the Probe Protocol}
\label{sec:instruments}
Three instruments estimate $\eps$, and they do not measure the same object.
(a)~The \emph{CRN best-response probe} runs paired deployments at
$\hRef\pm\delta$ under common random numbers. Its finite difference of best
responses estimates $m$, and $\eps=\hat m\gamma/\beta$ then inverts R1, so it
is a consistency check rather than an independent estimate. A signed $K$-step
variant measures $\mu(K)$ for R6. (b)~The \emph{optimal-transport} estimate
$\hat\eps=\Wone\big(D(h{+}\delta),D(h{-}\delta)\big)/2\delta$ on realized
toxic flow is the Wasserstein sensitivity of
\cite{perdomo2020performative} (Euclidean cost), which upper-bounds any mean
slope. It is computed via debiased log-domain Sinkhorn divergences
\cite{cuturi2013sinkhorn,feydy2019interpolating} with the blur set
\emph{scale-relatively} ($0.02\times$ pooled std, tuned against the exact 1-D
quantile $\Wone$; Sec.~\ref{sec:tuning}). (c)~The \emph{fitted informed-flow
curve} bins realized toxic notional against deployed spread, fits
\eqref{eq:flows}, and differentiates \cite{cont2014price}, a mean slope at
the realized state.

A pre-run audit showed three protocol facts that matter at first order. Probe at the operating spread rather than the
analytic fixed point. Keep collection jitter at $0.05$, since at $0.2$ the BR
probe inflates ${\sim}3\times$. And report medians and interquartile ranges
(IQR) over 8 seeds with the R4 robust bands at every grid point. We sweep the
gain $f$ rather than adversariality $\alpha$. High $\alpha$
drives the dealer to wide spreads where $e^{-c_t h}$ has decayed, so the
$\alpha$-response is non-monotone (measured hump $0.08\to1.83\to0.67$).

\subsection{Real-Data Calibration and Provenance}
\label{sec:data}
The simulator is calibrated per (rating $\times$ volatility regime) on a
public, verified panel. It combines ${\sim}36$ years of daily and ${\sim}70$
years of monthly series (CBOE VIX as the $\sigma$ proxy and regime
classifier, WTI, the Fed H.15 10-year, Shiller equity data, gold and CPI)
with the TRACE-derived corporate-bond factors of \cite{dickerson2023priced},
whose liquidity risk factor is the primary $\eps$ proxy, and monthly returns
for 212 real-CUSIP bonds. Half-spread proxies combine VIX-implied spreads
with the illiquidity
add-on of \cite{friewald2012illiquidity,bao2011illiquidity}, and
$\lambda(h)=Ae^{-kh}$ intensities are fit per cell by maximum likelihood.

On provenance, plainly: this is not trade-level TRACE. Dealer-side prints,
per-dealer inventories, and per-bond $(A,k)$ would require TRACE Enhanced. Only $(A,k,\sigma,h)$ are data-identified here, the
toxic channel is structurally scaled by documented ratios, and the
crisis-regime intensity fit is degenerate ($k=0$, $n=74$ days) and flagged
wherever it binds. Regime \emph{ordering} of the boundary is data-driven;
absolute critical gains are not. Calibrated configurations run in
per-\$100-par units, and every probe width, tolerance, and OT blur is
scale-relative by convention.

\subsection{Verification Layer}
\label{sec:verification}
Every load-bearing identity, inequality, and dynamical claim of R1--R6 is
re-derived numerically by 66 \emph{proof certificates}, run on both the raw
and the calibrated real-unit configurations: central differences of the
\emph{composed} best-response map against the assembled $-\eps\beta/\gamma$
with no shared code path, eigensolves against spectral formulas, Monte-Carlo
rate fits against claimed exponents, and the dynamics run rather than
assumed.

Building the layer caught real errors before the paper-grade run. It surfaced
a convention mismatch in which the 1-D frozen-gradient helpers omit the
inventory curvature, so their slopes are governed by $\gamma-P\lambda_q$. It
also falsified two first drafts: an ``always-stiffer'' claim for $\geff$,
replaced by the two-branch law of R6, and a mis-signed heavy-tail prediction
for the R4 radius. The logical skeletons of all six results are additionally
formalized in Lean~4 against mathlib \cite{moura2021lean,mathlib2020}. These
are reviewed statements whose compilation is pending a toolchain, so the
numerical certificates remain the verification of record. The full suite (152
tests, 11 experiments; Python, PyTorch, NumPy) runs CPU-only in ${\sim}25$
minutes, deterministic from (configuration, seed).

\section{Results}
\label{sec:results}

\subsection{Numerically Certified Identities}
All 66 certificates pass on the raw and the calibrated real-unit
configurations. The tightest is R1's cobweb linearization, with residual
$6.8{\times}10^{-2}$ against a tolerance of $8{\times}10^{-2}$; the R4 rate
fit sits at a third of its tolerance and every other family below
$10^{-3}$ of its own. The R2 demo-dynamics certificate pins absolute
raw-unit constants, so it runs on the raw configuration only.

\subsection{A Real-Data Fragility Index (1990--2026)}
\label{sec:fragility}
\begin{figure}[t]
\centering
\includegraphics[width=\columnwidth]{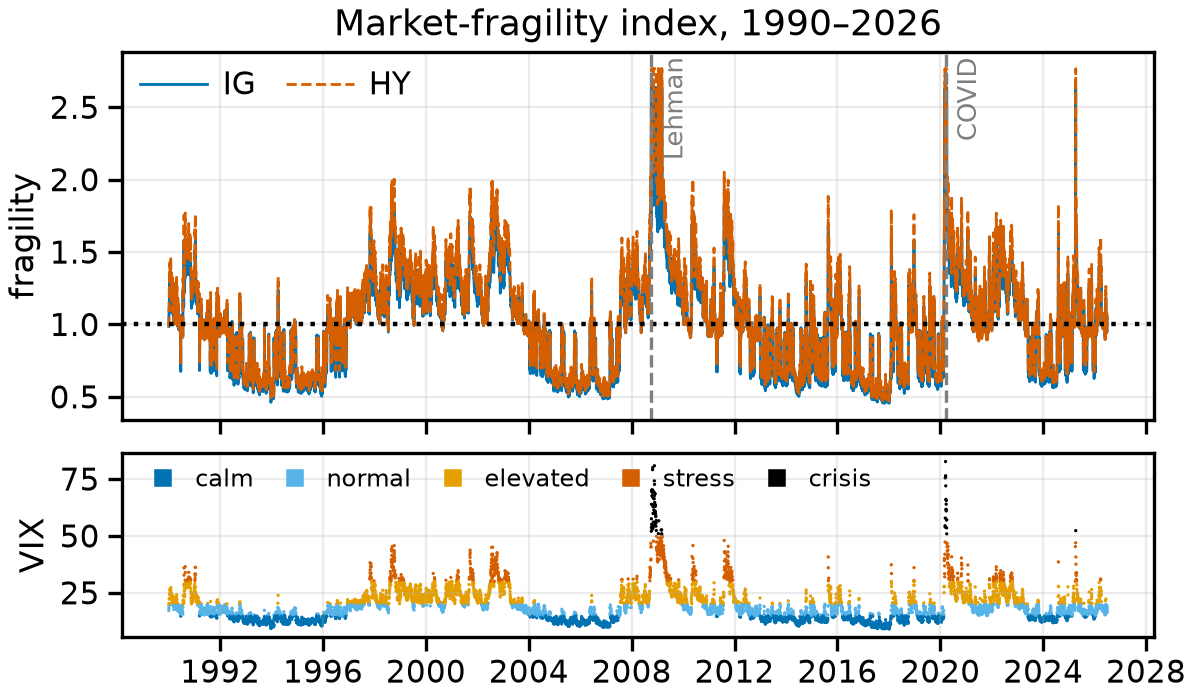}
\caption{The daily market-fragility index, 1990--2026: the R1 closed
forms evaluated on real data. Top:
$\mathrm{fragility}(t)=\mathrm{median}(\eps^*)/\eps^*(t)$ for IG (solid)
and HY (dashed), with the Lehman and COVID-freeze dates marked. Bottom:
the VIX spine, colored by volatility regime. The crisis plateau reflects
the degenerate crisis-cell fit and is flagged, not hidden.}
\label{fig:fragility}
\end{figure}

Because R1 is a closed form in observables, it can be evaluated on every
trading day of the panel. Figure~\ref{fig:fragility} plots the resulting
index, $\mathrm{fragility}(t)=\mathrm{median}(\eps^*)/\eps^*(t)$, and
Table~\ref{tab:fragility} gives the per-regime summary. First, the stability
headroom
$\eps^*=\gamma/\beta$ collapses
${\sim}4.4\times$ for investment grade (IG) and ${\sim}4.3\times$ for high
yield (HY) from calm to crisis. Second, HY sits more than $10\times$ below IG
in \emph{every} regime, which is a level effect rather than a crisis effect.
Third, the modulus at \emph{observed} spreads falls into crisis (IG
$0.85\to0.14$), so dealers widen faster than the toxic channel steepens.
The index therefore tracks shrinking \emph{headroom}, the smallest toxic
slope that would destabilize retraining. It does not track proximity to
$m=1$, which by the same table is greatest in calm markets.

The index saturates at a crisis plateau because of the degenerate crisis fit.
The GFC and the March-2020 COVID freeze are flagged at the same ceiling, so
intra-crisis ranking is explicitly not identified. Its full-sample median is
an ex-post normalizer, and the exercise shows that R1 can be evaluated on
real data, not that it is validated there.

\begin{table}[t]
\caption{Per-regime stability headroom $\eps^*=\gamma/\beta$ and modulus
at observed spreads, evaluated on the 1990--2026 panel.}
\label{tab:fragility}
\small
\begin{tabular}{@{}lrrrr@{}}
\toprule
regime & $\eps^*_{\mathrm{IG}}$ & $\eps^*_{\mathrm{HY}}$ &
$m_{\mathrm{IG}}(h_{\mathrm{obs}})$ & $m_{\mathrm{HY}}(h_{\mathrm{obs}})$\\
\midrule
calm     & 1207.7 & 93.5 & 0.847 & 0.592 \\
normal   &  739.2 & 61.3 & 0.870 & 0.563 \\
elevated &  594.4 & 45.9 & 0.520 & 0.367 \\
stress   &  477.3 & 35.6 & 0.223 & 0.163 \\
crisis   &  275.9 & 21.8 & 0.139 & 0.091 \\
\bottomrule
\end{tabular}
\end{table}

\subsection{Predict-Then-Verify: the Boundary Crossing}
\label{sec:sweep}
\begin{figure}[t]
\centering
\includegraphics[width=\columnwidth]{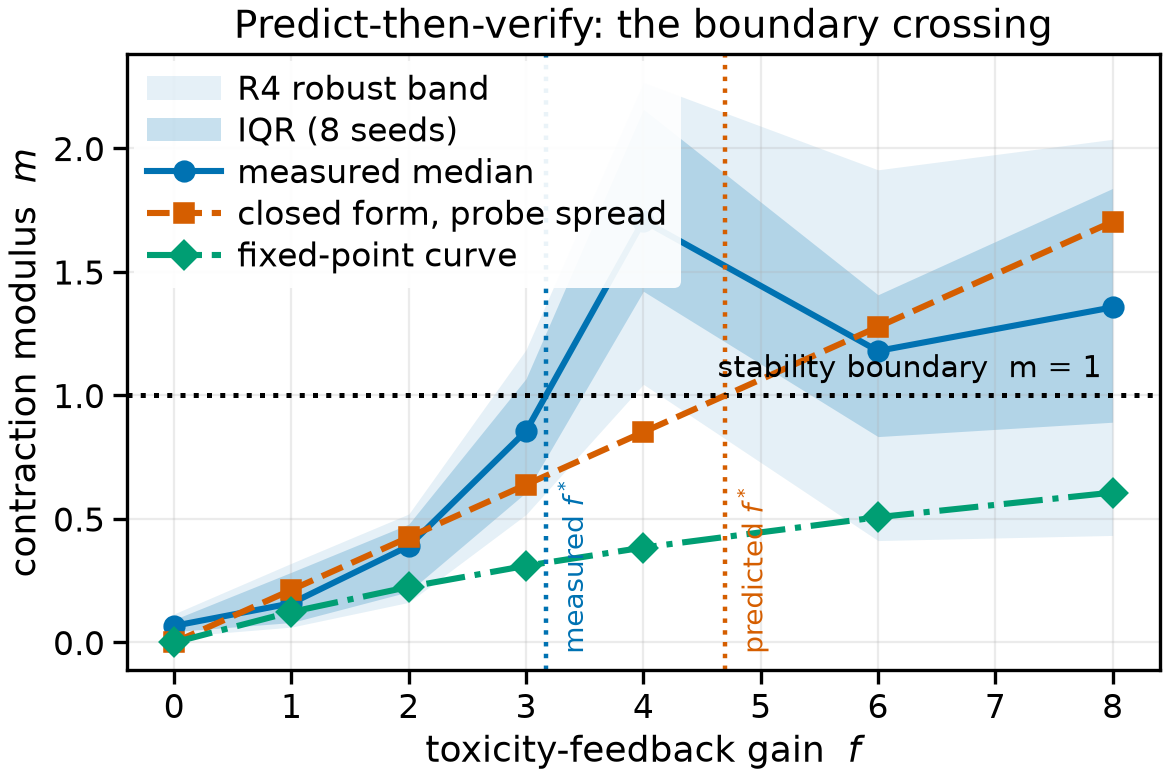}
\caption{Predict-then-verify phase diagram over the feedback gain $f$:
measured modulus (median, IQR band, and R4 robust band over 8 seeds)
against the a-priori closed form at the probe spread and the saturating
fixed-point curve, with the measured ($f^*\approx3.17$) and predicted
($4.70$) crossings marked.}
\label{fig:sweep}
\end{figure}

Figure~\ref{fig:sweep} is the headline falsification test. We sweep the gain
$f$ (7 values $\times$ 8 seeds), overlay the a-priori curve
\eqref{eq:modulus}, and measure the modulus with the CRN probe. In the
contracting regime the agreement is quantitative: measured median $0.390$
against predicted $0.426$ at $f=2$, an $8\%$ gap. The measured crossing,
interpolated between $f=3$ and the already-unstable $f=4$, sits at
$f^*\approx3.17$ against the a-priori $4.70$, so the a-priori form
over-certifies stability. Table~\ref{tab:triangulation} measures the
deployment's own flow inflating the liquidity field to $\rho\approx2.3$, and
the closed form evaluated at that \emph{realized} state predicts
$f^*\approx2.8$--$3.0$, just on the conservative side of the measurement.

The R4 certificates grade the grid exactly as the seed bands warrant: stable
through $f=2$, unstable at $f=4$, undecided where beyond-boundary readings
scatter. At $f=8$ the eight seeds span
$0.03$--$2.05$, because past the boundary the probe is a finite-difference
diagnostic rather than a local slope. The fixed-point curve itself saturates
at $0.61$, so the measured instability is a property of the retraining map at
the operating spread, exactly as R1 predicts.

Table~\ref{tab:triangulation} is taken at the default gain $f=5$, beyond the
crossing, and its readings span $0.51$--$4.58$, about $9\times$. The CRN slope inverts R1 from a beyond-boundary
probe reading. The Sinkhorn estimate is a $\Wone$ sensitivity and
upper-bounds any mean slope. The fitted curve is a mean slope like the closed
form, yet reads $2.3\times$ above it at the realized state. That residual is
the state-feedback channel a frozen-state closed form omits, so the closed
form is a \emph{lower anchor} for $\eps$ and, uncorrected, an optimistic
stability certificate. A desk should evaluate R1 at the realized state.

\begin{table}[t]
\caption{Three-way $\eps$ triangulation at the operating spread and default
gain $f=5$, against the closed form at the a-priori and at the realized
deployment state ($\rho=2.32$, $|g|=0.446$).}
\label{tab:triangulation}
\small
\begin{tabular}{@{}lrr@{}}
\toprule
instrument & $\hat\eps$ & vs.\ realized \\
\midrule
closed form, a-priori state ($\rho=1$) & 0.764 & $0.44\times$ \\
closed form, \textbf{realized state}   & 1.726 & $1.00$ \\
CRN best-response slope                & 0.508 & $0.29\times$ \\
Sinkhorn/$\Wone$                       & 4.578 & $2.65\times$ \\
Fitted informed-flow curve             & 4.011 & $2.32\times$ \\
\bottomrule
\end{tabular}
\end{table}

\subsection{Competition Amplifies Instability (R3)}
\label{sec:dealers}
We probe a \emph{genuine} $N$-dealer market sharing one informed pool rather
than an analytic shortcut; the environment reduces bit-for-bit to the
single-dealer market at $N=1$. CRN joint-mode probes measure the common-mode
slope against the predicted $\Neff m_1$ (Table~\ref{tab:dealers}).
Amplification is $1.74\times$ at $N{=}2$ and $3.16\times$ at $N{=}3$ against
predicted $2$ and $3$, within $13\%$ and $5\%$. The differential mode is dead
at full spillover ($(1-\kappa)m_1=0$), so the instability is purely the
synchronized systemic channel.

\begin{figure}[t]
\centering
\includegraphics[width=\columnwidth]{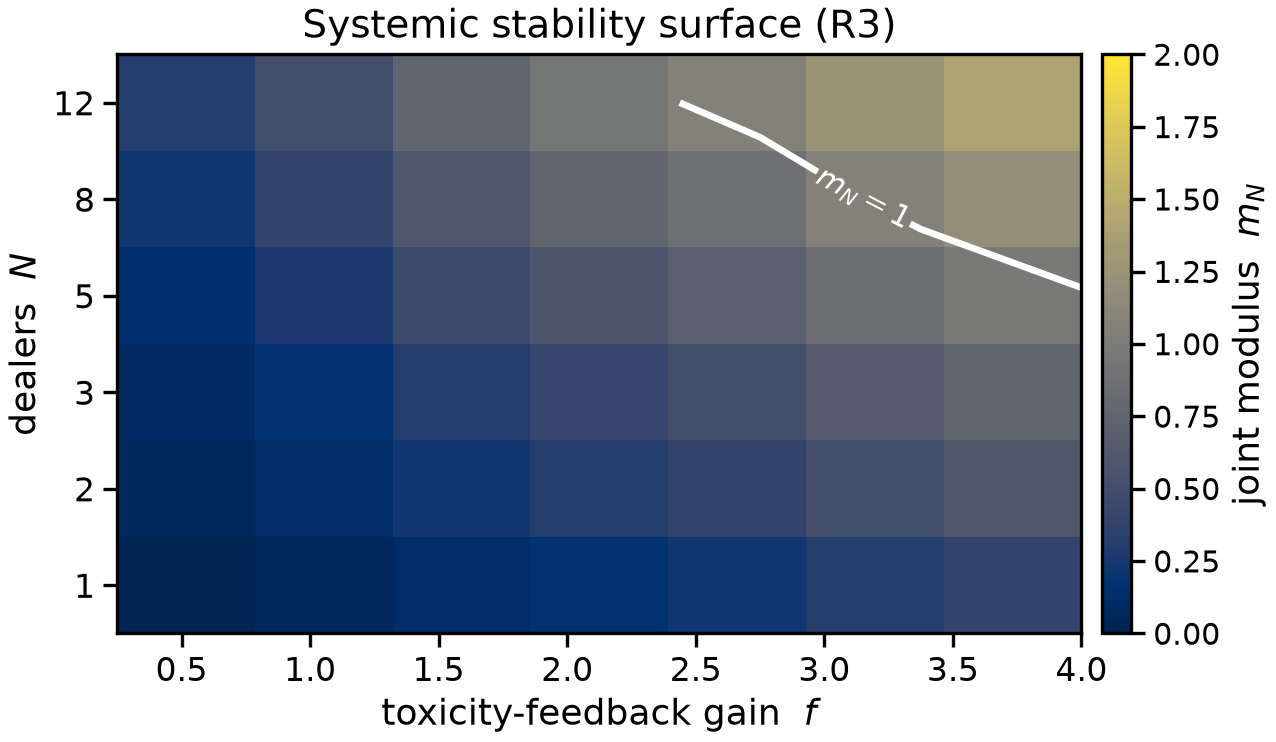}
\caption{The R3 systemic surface: joint modulus $m_N=\Neff\,m_1$ over the
$(f,N)$ grid at full spillover ($\kappa=1$), with the $m_N=1$ boundary
contour. Gains that are comfortably stable for a single dealer cross the
systemic boundary as the dealer count grows; the probes of
Table~\ref{tab:dealers} test the same $\Neff$ scaling on the simulated
market at a separate probe configuration.}
\label{fig:dealersurface}
\end{figure}

The analytic surface in Figure~\ref{fig:dealersurface} puts the critical
dealer count at $N_c\approx7.9$ for a single-dealer modulus $m_1\approx0.13$.
The probes run at a separate interior configuration with $m_1=0.786$, where
two dealers already cross, and the simulated three-dealer cobweb oscillates
against its cap. Each desk can satisfy its own stability
check while the market as a whole violates \eqref{eq:multidealer}, so
fragility has to be certified at the market level.

\begin{table}[t]
\caption{Measured common-mode retraining slope on the shared-pool market
versus the R3 prediction $\Neff m_1$ (full spillover, $\kappa=1$; one seed;
the $N=1$ row defines $m_1$).}
\label{tab:dealers}
\small
\begin{tabular}{@{}crrr@{}}
\toprule
$N$ & measured & predicted & differential mode \\
\midrule
1 & 0.786 & 0.786 & -- \\
2 & 1.369 & 1.571 & 0.0034 \\
3 & 2.480 & 2.357 & 0.0032 \\
\bottomrule
\end{tabular}
\end{table}

\subsection{Closing the Loop-Level Gap: Structural PerfGD (R2)}
\label{sec:gap}
\begin{figure*}[t]
\centering
\includegraphics[width=\textwidth]{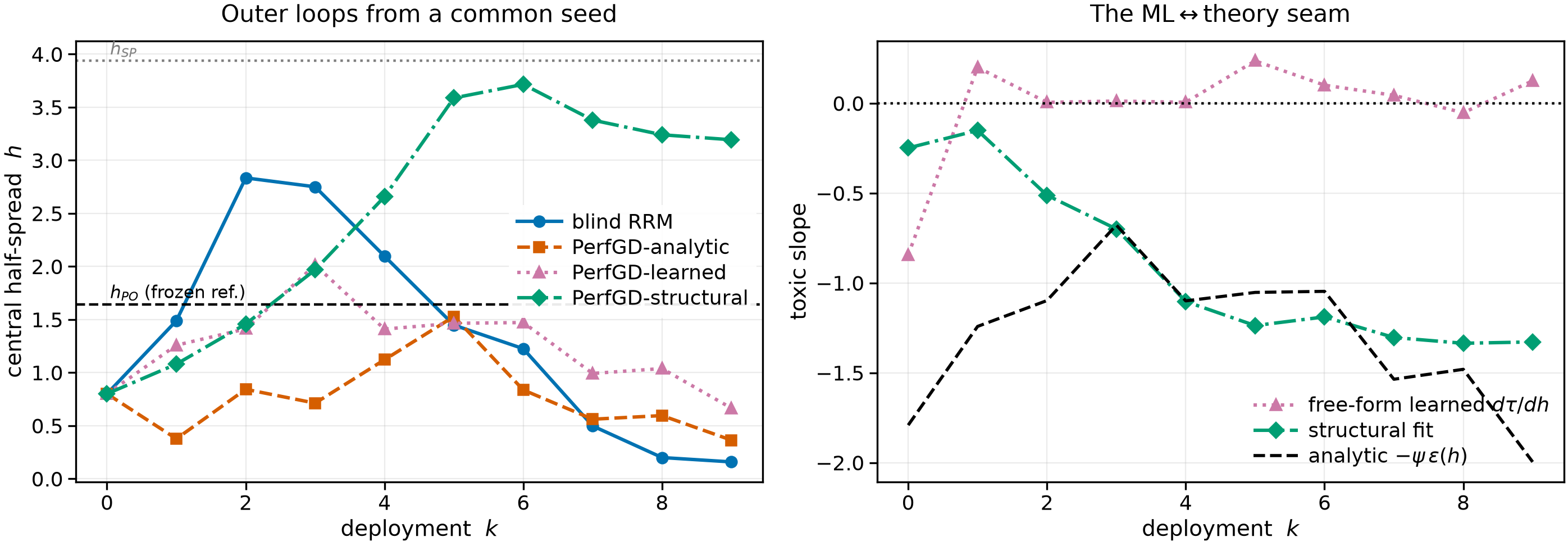}
\caption{Left: half-spread trajectories of the four learned retraining
loops from a common seed in the RRM-unstable regime (the frozen-reference
$\hPO$ and $\hSP$ lines are context, not the benchmark;
Sec.~\ref{sec:gap}). Blind RRM collapses; the structurally anchored loop
converges to the realized performative optimum. Right: the three-way
ML$\leftrightarrow$theory seam, showing the free-form learned, structurally
fitted, and analytic toxic slopes logged at every deployment. The
free-form slope flips sign and hovers near zero; the structural fit
tracks the analytic shape.}
\label{fig:loops}
\end{figure*}

The demo regime is genuinely RRM-unstable ($f=6$, slow toxic decay, cobweb
modulus $1.21$). There the blind cobweb provably diverges while the corrected
1-D ascent converges, in closed form. The open question is whether a
\emph{learned} loop can realize that correction. Figure~\ref{fig:loops} and
Table~\ref{tab:loops} answer it from a common seed.

Blind RRM collapses into the echo chamber, ending at half-spread $0.16$.
PerfGD-analytic and the free-form PerfGD-learned mode both fail to settle,
because each consumes the operator's implied objective gradient. The run's
seam diagnostics show why. The free-form learned toxic slope starts
right-signed ($-0.84$ against analytic $-1.79$), then flips sign and hovers
near zero. The network has ample capacity; what it lacks is identification
off the deployed regime. The structural mode never asks it for a derivative.
Fitting \eqref{eq:structfit} to its own deployment history, it climbs under
the step cap and \emph{converges to the realized performative optimum} at
$h=3.193$, with late steps falling $0.35\to0.02$. That is within $0.7\%$ of
its own running estimate $\hat h_{\mathrm{PO}}=3.172$, and its fitted slope
tracks the analytic shape ($-1.33$ against $-2.00$ at the final iterate)
while strengthening as the fit window widens.

\emph{Benchmark honesty.} The frozen-reference closed-form optimum
($\hPO=1.641$) is \emph{not} the target. In this high-intensity regime the
realized market differs from the frozen closed forms at first order through
channels they omit by construction: informed-flow saturation, the
$\rho\approx2.3$ liquidity inflation of Sec.~\ref{sec:sweep}, and severity
drift. The loop is therefore verified against \emph{independent structural
fits} on fresh controlled deployments spanning the operating range. Against
those fits the settle point (i) zeroes their corrected gradient (residual
$<30\%$ of its start value), (ii) sits strictly inside their blind stable
point ($<0.85\times\hat h_{\mathrm{SP}}$, the realized echo-chamber gap,
closed), and (iii) brackets their independent optimum estimate. We claim only
this much: un-blinding is proven in closed form, and the structurally
anchored learned loop realizes it at loop level against the realized-market
benchmark, while the free-form correction remains a documented negative
result.

\begin{table}[t]
\caption{Loop outcomes in the RRM-unstable demo regime (10 deployments,
common seed; convergence verified against independent structural fits).}
\label{tab:loops}
\small
\begin{tabular}{@{}lrl@{}}
\toprule
mode & final $h$ & outcome \\
\midrule
blind RRM          & 0.16 & echo-chamber collapse \\
PerfGD-analytic    & 0.36 & no settle (operator gradient) \\
PerfGD-learned     & 0.66 & no settle (negative result) \\
\textbf{PerfGD-structural} & \textbf{3.19} & \textbf{converges to realized optimum} \\
\bottomrule
\end{tabular}
\end{table}

\subsection{Lazy Deployment Stabilizes an Unstable Market (R6)}
\label{sec:lazy}
\begin{figure*}[t]
\centering
\includegraphics[width=\textwidth]{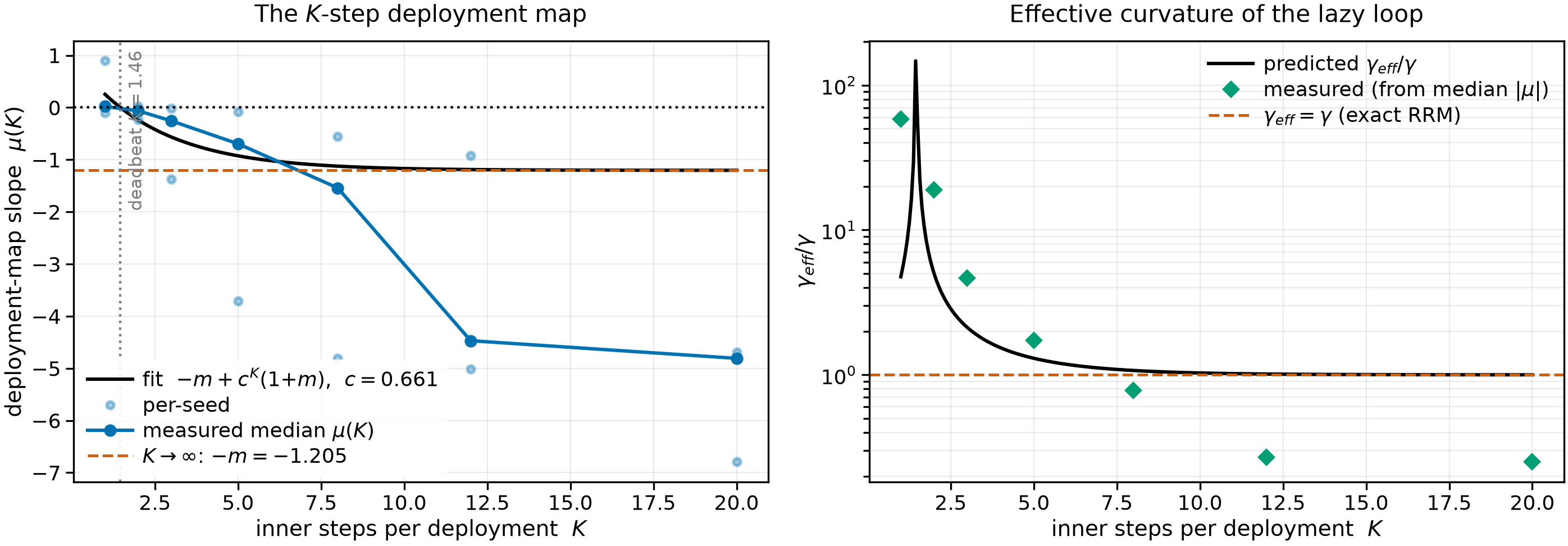}
\caption{Lazy deployment at a beyond-boundary configuration
($\hat m=1.205$). Left: the measured signed $K$-step map (medians over
seeds; faint dots are per-seed readings) against the one-parameter fit
$\mu(K)=-m+c^K(1+m)$ with $c=0.661$, its $K\to\infty$ asymptote $-m$, and
the predicted deadbeat cadence. Right: the implied effective curvature
$\geff(K)/\gamma$, predicted against measured, with the stiff branch
decaying through the exact-RRM line between $K=5$ and $K=8$.}
\label{fig:lazy}
\end{figure*}

At a beyond-boundary configuration ($f=5$, exact-BR anchor $\hat m=1.205>1$
measured on the same seeds), the signed CRN $K$-step probe traces $\mu(K)$
for $K\in\{1,2,3,5,8,12,20\}$ over 3 seeds (Figure~\ref{fig:lazy}). A
one-parameter fit of \eqref{eq:lazy} gives inner contraction $c=0.661$, and
with that one fitted parameter both predicted cadences match. The deadbeat sign flip is predicted at $\Kdb=1.46$, and
measured medians run $+0.021$ at $K{=}1$ and $-0.064$ at $K{=}2$. The
stability-window exit is predicted at $\Kmax=5.75$, and measured $|\mu|$ is
$0.695$ at $K{=}5$ (inside) and $1.541$ at $K{=}8$ (outside).

Laziness therefore keeps this RRM-unstable market measurably stable for
$K\lesssim6$ and inherits the cobweb's divergence beyond, which makes
retraining \emph{cadence} a stability control. The measured $\geff/\gamma$
falls monotonically through $1$ between $K=5$ and $K=8$, the stiff branch of
the two-branch law. The law holds only for $K\le8$. At $K=12$ and $20$ the
medians ($-4.47$, $-4.81$) fall far below the asymptote $-m$, which
\eqref{eq:lazy} cannot produce: at that depth the probe compounds the
beyond-boundary bifurcation. The clean tight-fit
demonstration lives in the contracting regime, where measured
$+0.78/+0.70/+0.37$ meet fitted $+0.88/+0.68/+0.36$ at $K=1/3/8$.

\subsection{Scaling and Estimator Tuning}
\label{sec:tuning}

\begin{figure*}[t]
\centering
\includegraphics[width=\textwidth]{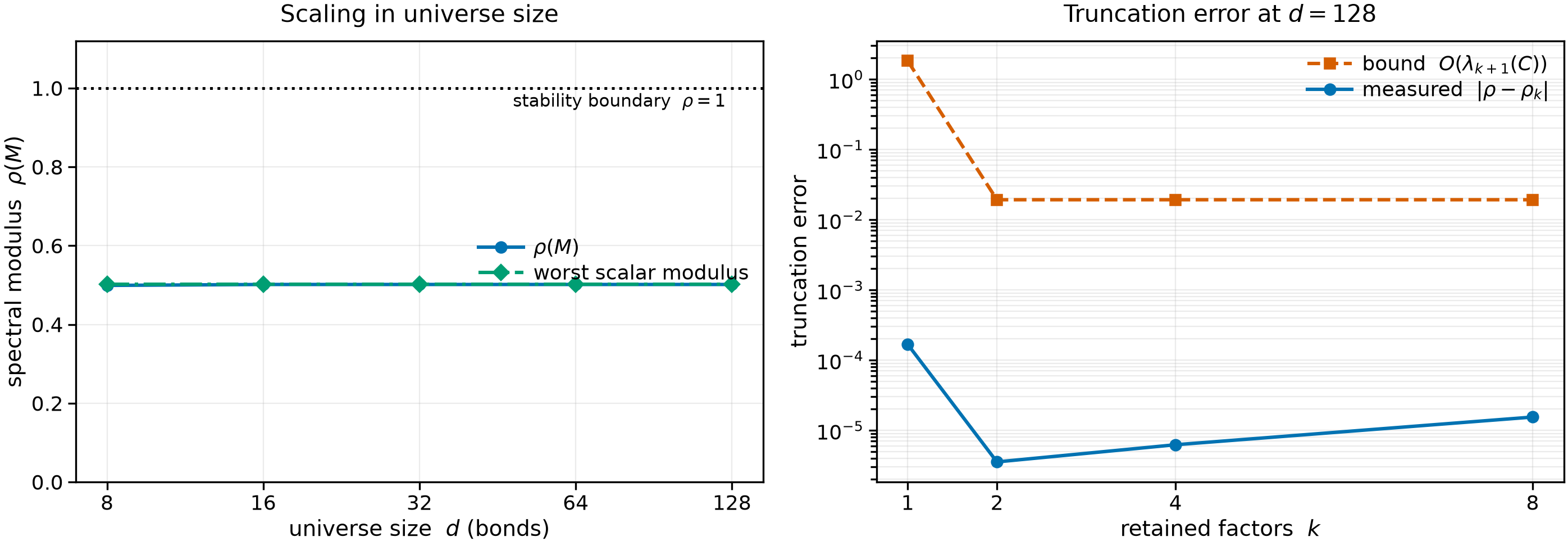}
\caption{Factor-model scaling (R5) with data-calibrated per-bond
dispersion. Left: $\rho(M)$ is flat in universe size from $d=8$ to $128$
bonds and pinned to the worst scalar modulus, far below the $\rho=1$
boundary (dotted). Right: the truncation bound $O(\lambda_{k+1}(C))$
against the measured error $|\rho-\rho_k|$ at $d=128$, showing three to four
orders of magnitude of slack at every retained-factor count $k$.}
\label{fig:universe}
\end{figure*}

With per-bond volatilities dispersed at the data-calibrated
cross-sectional coefficient of variation, $\rho(M)$ stays at
${\approx}0.50$ from $d=8$ to $128$ bonds (Figure~\ref{fig:universe}),
pinned to the worst scalar modulus, so on this calibration the fragile mode
is idiosyncratic (R5); the truncation bound holds with 3--4 orders of
magnitude of slack. Against the exact 1-D quantile $\Wone$, the Sinkhorn
bias is smallest at a scale-relative blur of $0.02\times$ std ($6.1\%$
ground-truth bias, $1.2\%$ on the CRN samples). The $z\,s$ robust radius
(R4) over-covers ($0.99$--$1.00$ at $95\%$ nominal) except under
railed-probe contamination ($0.938$).

\section{Conclusion}
This paper proposes REFLEX, which gives a quoting desk's retraining loop a
stability condition the desk can evaluate. We compute the cobweb's
$(\eps,\beta,\gamma)$ from microstructure primitives, extend the boundary in
closed form to correction, competition, finite samples, dimension, and
cadence, test each extension predict-then-verify against a learned loop,
evaluate it as a daily fragility index on 36 years of real market data, and
certify the identities numerically. Two of the findings should carry past this model.
Competition among retraining dealers amplifies instability by the effective
dealer count, which makes certification of algorithmic liquidity provision a
market-level rather than a desk-level problem. And a learned corrected loop
stabilizes where blind retraining collapses only when its response model is
anchored to market structure. The free-form alternative fails with the same
data and more capacity, so a desk should spend its modelling budget on
structural anchoring rather than on network capacity.

These results hold under idealizing assumptions. First, the calibration is
proxy-level rather than trade-level TRACE: VIX-implied
spreads, no per-dealer inventories, a structurally scaled toxic channel, and
a degenerate crisis cell ($k=0$). Regime ordering is therefore data-driven
while absolute critical gains are not, and TRACE Enhanced calibration removes
that limit. Second, at default-like constants the self-consistent fixed point
never destabilizes because dealers widen defensively, so measured crossings
describe the retraining map at the operating spread. Beyond the boundary,
probe readings are finite-difference diagnostics with seed-level bifurcation,
so medians with robust bands carry the results, and moduli are comparable
within a protocol rather than across protocols. Third, the structural loop's
optimum is the \emph{realized} one, benchmarked against independent
structural fits rather than the frozen-reference closed form, with the gap
attributed to saturation, liquidity inflation, and severity drift. Its
convergence is to the noise ball of an estimated gradient, and the free-form
learned correction remains a negative result by design. Finally, the Lean~4
skeletons are reviewed but not yet compiled, so the 66 numerical certificates
are the verification of record. Future work: trade-level calibration,
inventory-state loop dynamics under state-dependent performativity, richer
policy classes, and compiling the formal layer.

\section*{Reproducibility}
\label{sec:repro}
The full framework is openly available:\footnote{%
\url{https://github.com/vignesh-nagarajan-vn/REFLEX/}}
the six derivation documents D1--D6, the simulator, the operator, all four
loop modes, the estimator suite, the calibration pipeline with its data
catalogue, the 66-certificate verification layer,
the Lean sources, and the illustrated reports of both paper-grade runs.
All results in this paper are regenerated by one command
(\code{run\_all --profile full}, ${\sim}25$ CPU-minutes, deterministic from
configuration and seed).

\bibliographystyle{ACM-Reference-Format}
\bibliography{icaif26-177}

\end{document}